\documentclass{article}
\usepackage{ijcai22}

\usepackage{times}
\usepackage{soul}
\usepackage{url}
\usepackage[hidelinks]{hyperref}
\usepackage[utf8]{inputenc}
\usepackage[small]{caption}
\usepackage{graphicx}
\usepackage{amsmath}
\usepackage{amsthm}
\usepackage{booktabs}
\usepackage{algorithm}
\usepackage{algorithmic}
\usepackage{multirow}
\title{Personalized Federated Learning With Graph}

\author{
Fengwen Chen$^1$\and
Guodong Long$^1$\and
Zonghan Wu$^1$\and
Tianyi Zhou$^{2,3}$ \And
Jing Jiang$^1$\footnote{Contact Author}\\
\affiliations
$^1$Australian Artificial Intelligence Institute, FEIT, University of Technology Sydney\\
$^2$University of Washington, Seattle\\
$^3$University of Maryland, College Park\\
\emails
\{Fengwen.Chen, Zonghan.Wu-3\}@student.uts.edu.au, \\
\{Guodong.Long, Jing.Jiang \}@uts.edu.au,
tianyizh@uw.edu
}

\begin{document}

\maketitle

\begin{abstract}
Knowledge sharing and model personalization are two key components in the conceptual framework of personalized federated learning (PFL). Existing PFL methods focus on proposing new model personalization mechanisms while simply implementing knowledge sharing by aggregating models from all clients, regardless of their relation graph. This paper aims to enhance the knowledge-sharing process in PFL by leveraging the graph-based structural information among clients. We propose a novel structured federated learning (SFL) framework to learn both the global and personalized models simultaneously using client-wise relation graphs and clients' private data. We cast SFL with graph into a novel optimization problem that can model the client-wise complex relations and graph-based structural topology by a unified framework. Moreover, in addition to using an existing relation graph, SFL could be expanded to learn the hidden relations among clients. Experiments on traffic and image benchmark datasets can demonstrate the effectiveness of the proposed method. All implementation codes are available on Github\footnote{https://github.com/dawenzi098/SFL-Structural-Federated-Learning}.
\end{abstract}

\section{Introduction}

Since Federated Learning (FL)\cite{mcmahan2017communication} was first proposed in 2017, it has evolved into a new-generation collaborative machine learning framework with applications in a range of scenarios, including Google's Gboard on Android \cite{mcmahan2017communication}, Apple's siri \cite{apple2017federated}, Computer Visions \cite{luo2019real,jallepalli2021federated,he2021fedcv}, Smart Cities \cite{zheng2022applications}, Finance \cite{long2020federated} and Healthcare \cite{rieke2020future,xu2021federated,long2022federated}. Various FL tools and packages have been developed and open-sourced by hi-tech companies, such as Google\footnote{https://www.tensorflow.org/federated}, NVIDIA\footnote{https://nvidia.github.io/NVFlare}, Intel\footnote{https://github.com/intel/openfl},
Amazon \footnote{https://aws.amazon.com/blogs/architecture/applying-federated-learning-for-ml-at-the-edge/}, Baidu\footnote{https://github.com/PaddlePaddle/PaddleFL}, and Webank\footnote{https://github.com/FederatedAI/FATE}.

The vanilla FL method, known as FedAvg \cite{mcmahan2017communication}, is derived from a distributed machine learning framework before it is applied to a large-scale mobile service system. In particular, it aims to train a single shared model at the server by aggregating the smartphones' local model, trained with its own data. Thus, the end-user's private data in each smartphone is not uploaded to the cloud server. FedAvg first proposed the non-IID challenge, a key feature of FL.
To tackle the non-IID challenge in FL, some works \cite{li2018federated,reddi2020adaptive} focus on training a single robust model at the server. However,  other methods aim to learn multiple global or centric models \cite{ghosh2020efficient,mansour2020three,xie2021multi}, where each model serves a cluster of clients whose data distribution is the same or similar. More recently, personalized FL methods \cite{li2021ditto,deng2020adaptive,tan2022toward,fallah2020personalized} are proposed to learn many client-specific personalized models using the global model as the component of knowledge sharing. Therefore, the objective of FL research has been changed from learning server-based models to client-specific models.

\begin{figure}[tbp]
\centering
\includegraphics[width=\linewidth]{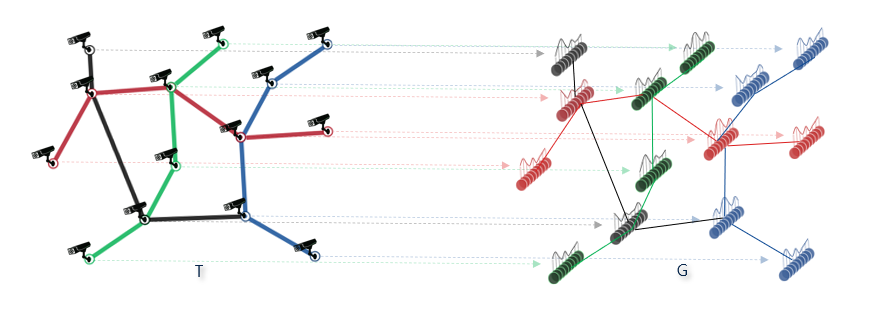}
\caption{In a smart city, each devices deployed on the road can collect data and make real-time decision without waiting for the response of cloud servers. Each device needs to make intelligent decision based on the collected road conditions and nearby devices.}
\setlength {\belowcaptionskip} {-1.5em}
\label{fig:intro}
\end{figure}

To learn client-specific personalized models in the FL setting, knowledge sharing and personalization of local models are two key components. However, existing personalized FL focuses on improving components of model personalization while implementing the component of knowledge sharing by simply aggregating all local models. This kind of implementation overlooks the graph relationship across clients with non-IID data. Moreover, if the degree of non-IID is very high, aggregation across all clients will produce a low-quality global model that will eventually impact the performance of model personalization. From an application perspective, graph relationship have many real-world applications, such as traffic sensors with road maps (Figure \ref{fig:intro}), devices in a smart home, mobile APPs using users' social networks, and fraud detection based on the mutual interactions of users. Adopting these types of relation graphs will enhance the performance of FL.

This paper proposes a novel \textbf{structured federated learning (SFL)} that aims to leverage the relation graph among clients to enhance personalized FL. In particular, we design a fine-grained model aggregation mechanism to leverage each client's neighbors' local models. Specifically, a relation graph will be stored in the server, and then the client-centric model aggregation will be conducted along the relation graph's structure. To simplify implementation, we propose to use the Graph Convolutional Network (GCN) \cite{kipf2016semi} to implement the model aggregation function; therefore, the proposed solution is easy-to-implement by integrating FL and GCN. In addition, we formulate the problem in a unified optimization framework to include both personalized FL and graph-based model aggregation. \textbf{Contributions} of this paper are summarized as follows.

 \begin{itemize}
    \item We are the first to propose a new federated setting by considering a relation graph among clients. Moreover, the proposed GCN-based model aggregation mechanism is a new and easy-to-implement idea for FL;
    \item The research problem has been formulated to a unified optimization framework that can learn optimal personalized models while leveraging the graph;
    \item The proposed method has been expanded to learn the hidden relations among clients, and the conceptual framework can be extended to integrate with other model personalization techniques;
    \item Experiments on both the image and traffic datasets have demonstrated the effectiveness of the proposed method. 
\end{itemize}

\section{Related Work}
\subsection{Federated learning with non-IID}
The vanilla FL method, FedAvg \cite{mcmahan2017communication}, has been suffering from the non-IID challenge where each client's local data distribution is varied \cite{kairouz2021advances}. To tackle this challenge,
\cite{li2019feddane} proposed FedDANE by adapting the DANE to a federated setting. In particular, FedDANE is a federated Newton-type optimization method. \cite{li2018federated} proposed FedProx for the generalization and re-parameterization of FedAvg. It adds a proximal term to clients' local objective functions by constraining the parameter-based distance between the global model and local model. \cite{reddi2020adaptive} proposes to use adaptive learning rates to FL clients and \cite{jiang2020decentralized} conduct attention-based adaptive weighting to aggregate clients' models. \cite{li2019convergence} studies the convergence of the FedAvg on non-IID scenarios. In summary, the early-stage FL research focuses on learning a robust global model that usually assumes the non-IID scenario is relatively mild. In recent times, research on personalized FL (Sec \ref{sec:PFL}) has attracted broad interest in tackling severe non-IID scenarios.


\begin{figure*}[tbp]
\centering
\setlength{\abovecaptionskip}{0.cm}
\includegraphics[width=0.75\textwidth]{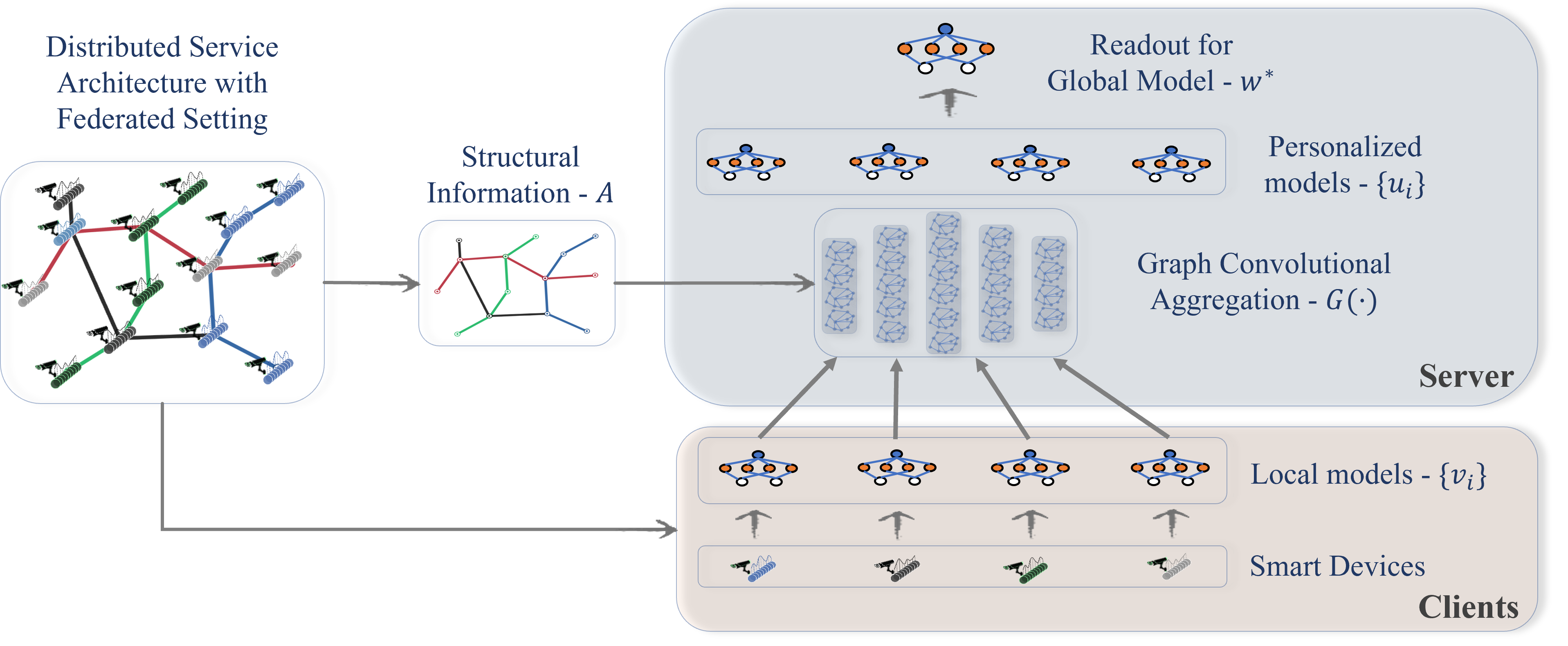}
\vspace{0.5em}
\caption{The overview of structured federated learning (SFL). A GCN module is used in the server to generate personalized client-specific models $\{u_i\}$ by aggregating the collected local model parameters $\{v_i\}$ according to the graph structure $A$ of all clients.}
\label{fig:overview}
\vspace{-1em}
\end{figure*}


\subsection{Personalized federated learning} \label{sec:PFL}
This section will discuss two major FL solutions to tackle severe non-IID scenarios. \textbf{Group-wise PFL}, which is also named clustered FL, assumes the clients can be clustered to different groups with severe non-IID across inter-group clients and mild non-IID across intra-group clients. Hence, clustered FL can be categorized according to different clustering methods and distance measurements. Kmeans-based clustered FL \cite{xie2021multi} and \cite{mansour2020three,ghosh2020efficient} measured the distance using model parameters and accuracy respectively. Hierarchical clustering \cite{briggs2020federated} has been applied to FL. Furthermore, \cite{ma2022convergence} proposed a general form to model the clustered FL problem into a bi-level optimization framework, then leveraged the important contributions among clients to form a weighted client-based clustered FL framework.

\textbf{Client-wise PFL} that usually assumes each client's data distribution is different from others; thus, each client should have a personalized model on their device. 
In general, a simple PFL method could train a global model in FedAvg, then conduct a few steps of fine-tuning on each client \cite{cheng2021fine}. In this framework, knowledge sharing is model aggregation, and model personalization is local fine-tuning. Per-FedAvg \cite{fallah2020personalized} considered fine-tune as a regularization term on the learning objective function of the global model. Ditto \cite{li2021ditto} was proposed as a bi-level optimization framework for PFL while considering a regularization term to constrain the distance between the local model and global model. Investigations by \cite{shamsian2021personalized,chen2018federated} that aim to train a global hyper-network or meta-learner instead of a global model before sending it to clients for local optimization. SCAFFOLD \cite{karimireddy2020scaffold} proposes to learn personalized control variate that correct the local model accordingly. Layer-wise personalization \cite{arivazhagan2019federated,liang2020think} and Representation-wise personalization \cite{tan2021fedproto} are two simple but effective solution of PFL.

\subsection{Learning with Structural Information}
Learning on structural data with GCNs \cite{kipf2016semi} or graph neural networks \cite{wu2020comprehensive} are ubiquitous in many fields for tasks such as , node classification \cite{pan2016tri}, link prediction\cite{wang2017predictive}, node clustering\cite{wang2017mgae}, and graph classification \cite{pan2016joint}. GCNs capture relationships between concepts (also called nodes) in a graph using the k-hop aggregation mechanism. Thus, a weighted hop in GCN can capture a more complex relationship in the graph \cite{chen2019dagcn}. Recently, GraphFL\cite{wang2020graphfl} has been proposed to explore the graph learning problem in a federated setting. In particular, each client trains a graph neural networks (e.g., GCNs) for learning the structured data locally, then shares knowledge via model aggregation at the server.

\section{Problem Formulation}
Given $N$ participants in an FL system, each one has a local dataset $D_i$ which is drawn from a distribution $P_i$. Given the non-IID setting, we usually assume all $P_i$ are distinct from each other. An adjacency matrix $A \in \lbrace 0,1 \rbrace^{N \times N}$ represents the topological relationship across participants. 
In general, a FL system is to solve below optimal objective. 
\begin{equation} \label{eq:global-loss-FL}
    \min_w G(F_1(w), ... F_N(w))
\end{equation}
where $F_i(W)$ is the supervised loss of the $i$-th client that has a locally stored dataset $D_i$, and all clients using the same global model $M$ parameterized by $w$. The $G(.)$ is a function that aggregates the local objectives. For example, in FedAvg \cite{mcmahan2017communication}, $G(.)$ is a weighted average of local lossess using the size of local dataset, ie.e., $\sum |D_i|/\sum_j |D_j|$.

In general, a personalized FL system is usually modeled as a bi-level optimization problem as below.
\begin{equation} \label{eq:global-loss-PFL}
  \begin{aligned} 
    &\min_{\{v_1...v_N\}} & h_i(v_i; w^*) := F_i(v_i) + \lambda R(v_i, w^*) \\
    &s.t. & w^* \in \arg\min_{w} G(F_1(w),...,F_N(w))
  \end{aligned}
\end{equation}
where each client has a unique personalized model $M_i$ parameterized by $v_i$, and $w^*$ is an optimal global model to minimize the loss as mentioned in the Eq. \ref{eq:global-loss-FL}. $R$ is the regularization term to control model updates on clients, for example, \cite{li2021ditto} propose a L2 term $\frac{1}{2} ||v_i - w^*||^2$ to constraint the local updating won't be far away to the global model.

To find the optimal solution for the loss Eq. \ref{eq:global-loss-PFL}, existing personalized FL methods will take various forms, such as fine-tuning \cite{cheng2021fine}, meta-training \cite{fallah2020personalized}, and partial parameter-sharing \cite{liang2020think}. Our proposed structured federated learning is a new solution to leverage both structural information and model parameters for personalized FL.

\section{Structured Federated Learning}

Our proposed structured FL will formulate the below bi-level optimization problem. 
\begin{equation} \label{eq:global-loss-SFL}
  \begin{aligned}
    &\min_{\{v_1...v_N\}} & \sum_{i=1}^N \left(F_i(v_i) + \lambda [R(v_i, w^*) + R(v_i, u_i)]\right)\\
    &s.t. &w^* \in \arg\min_{w} G(F_1(w),...,F_N(w)) \\
    &     &u_i \in \arg\min_{u_i}  \sum_{j\in\mathcal N(i)}  A_{j,i}S(u_i,u_j),
  \end{aligned}
\end{equation}
where the $A_{i,j} \in \{0,1\}$ is from the client-wise relation graph's adjacent matrix, and the $S(u_i,u_j)$ is to measure the distance, e.g. Euclidean distance, between the $i$-th client and its neighbor j, using their models' parameters. 

In many real applications, the adjacency matrix $A$ may not always exist. Thus, it needs to be learned. For this case, we can formulate the optimization problem as below.
\begin{equation} \label{eq:global-loss-SFLA}
  \begin{aligned}
    &\min_{\{v_1...v_N\};A} & \sum_{i=1}^N \left(F_i(v_i) + \lambda\mathcal{R}(v_i, w^*, u_i)\right) + \gamma \mathcal{G}(A)\\
    &s.t. & w^* \in \arg\min_{w} G(F_1(w),...,F_N(w)) \\
    &     & u_i \in \arg\min_{u_i}  \sum_{j\in\mathcal N(i)}  A_{j,i}S(u_i,u_j)
  \end{aligned}
\end{equation}
where $\mathcal{R}(.)=R(v_i, w^*) + R(v_i, u_i)$ is an abbreviation of two regularization terms.  $\mathcal{G}(.)$ is a regularization term for the learned graph with adjacent matrix $A$ (we will also call a graph $A$) which is expected to be sparse and able to preserve proximity relationship among clients. There are various ways to measure the proximity between two clients: for example, distance of model parameters, local accuracy using the same model, and external descriptive features.

\subsection{Optimization}
To solve the optimization problem in Eq. \ref{eq:global-loss-SFL}, we could conduct the below steps. First, we update the $v_i^*$ by solving the local loss $F_i(v_i)$ with two regularization terms: distance between local model and gradient-based aggregated global model $R(v_i, w^*)$, and distance between local model and graph-based aggregated personalized model $R(v_i, u_i)$. Then, we conduct model aggregation at the server to update $w$ and $\{u_i\}_i^N$. In particular, we can use a Graph Convolution Network (GCN) \cite{kipf2016semi} to implement the graph-based model aggregation by constructing the graph: N clients represent the node in the graph, a pre-defined adjacency matrix $A$, and each node's attribute $u_i$ is initialized by its local model $v_i$. The GCN module will automatically update each node's model parameters $u_i$ by aggregating its neighbors' model in the graph. It will satisfy the second constraint in Eq. \ref{eq:global-loss-PFL}. Moreover, the global model $w$ will be updated by aggregating all personalized models $u_i$ which is to satisfy the first constraint in Eq. \ref{eq:global-loss-SFL}. This gradient-based model aggregation across all clients is equivalent to the read-out operator in the GCN.

To solve the optimization problem in Eq. \ref{eq:global-loss-SFLA}, we can add a structure-learning step in the aforementioned optimization steps for Eq. \ref{eq:global-loss-SFL}. In particular, we will design a graph encoder to minimize three regularization terms of Eq. \ref{eq:global-loss-SFLA}, as below. 
\begin{equation}
    \min_{A} \sum_{i=1}^N \left(\lambda [R(v_i, w^*) + R(v_i, u_i)] + \gamma \mathcal{G}(A) \right)
\end{equation}

We can construct the relation graph $A$ using the learned representation of nodes. We can also define a fully connected graph with weighted edges. The GCN will not only learn representation but also learn the structure by adjusting the weights of edges.

\begin {algorithm}[htb]
\caption {\small Structural Federated Learning - Server.}
\label{alg:alg}
\begin {algorithmic}[1]
\STATE Initialize $\lambda_0, \eta, A, \{v_i^{(0)}\}_{i=1}^N \leftarrow v $ 
\FOR{each communication round t = 0, 1, ..., T}
    \STATE $\lambda = 1[t>0] \times \lambda_0$
    \smallskip 
    \STATE \underline{\textit{Local updating:}}
    \smallskip 
    \FOR{each client i = 1, 2, ...., N in parallel}
        \STATE Update $v_i$ for $s$ local steps:
        \smallskip 
        \smallskip 
        \STATE $v_i^{(t)}\leftarrow v_i^{(t)} - \newline ~~~~~~~~~~\eta\nabla\left(F_i(v_i^{(t)}) + \lambda [R(v_i^{(t)}, w^{(t)}) + R(v_i^{(t)}, u_i^{(t)})]\right)$
        \STATE $v_i^{(t+1)}\leftarrow v_i^{(t)}$
    \ENDFOR
    \smallskip
    \smallskip 
    \STATE \underline{\textit{Structure-based aggregating:}} 
    \smallskip
    \STATE $\{u_i^{(t+1)}\}_{i=1}^N \leftarrow \{v_i^{(t)}\}_{i=1}^N $
    \STATE Update $u_i^{(t+1)}$ for m steps of $GCN(A,~\{u_i^{(t+1)}\}_{i=1}^{N})$ \\
    \STATE $w^{(t+1)} \leftarrow GCN\_readout(\{u_i^{(t+1)}\}_{i=1}^N)$
    \smallskip 
    \smallskip 
    \STATE \underline{\textit{(Optional) Structure learning:}}
    \smallskip
    \STATE $A\leftarrow Structure\_learn(\{v_i^{(t+1)}, u_i^{(t+1)},w^{(t+1)}\}_{i=1}^N)$.  \\
\ENDFOR
\end {algorithmic}
\end {algorithm}
\subsection{Algorithm}
The aforementioned optimization procedure has been implemented in Algorithm \ref{alg:alg}. The optimization goal will be achieved iterative through multiple communication rounds between the server and clients. In each communication round, we will have two steps to solve the bi-level optimization problem. First, we update the local model $v_i$ by conducting local model training with supervised loss and regularization terms. Second, we conduct model aggregation at the server using GCN. In the case that A is not present, we will add an optional step for structure learning.

\begin{table*}[tpb]
\resizebox{2.0\columnwidth}{!}{
\begin{tabular}{c|ccc|ccc|ccc|ccc}
    & \multicolumn{3}{c|}{METR-LA} & \multicolumn{3}{c|}{PEMS-BAY} & \multicolumn{3}{c|}{PEMS-D4} & \multicolumn{3}{c}{PEMS-D8} \\ \hline
    & MAE  & MAPE & RMSE  & MAE     & MAPE     & RMSE     & MAE     & MAPE     & RMSE    & MAE     & MAPE    & RMSE    \\
    
FedAvg  & 7.03  & 21.63    &  10.81  & 3.62    & 10.65    & 7.26     & 44.96   & 30.03    & 59.97   & 36.76   &  21.04  &  49.14  \\

FedAtt  & 6.89  & 23.54    & 10.55   & 3.26    & 5.50    & 6.41     & 45.53   & 30.15    & 60.68   & 35.80   &  23.27  & 47.75   \\  

SFL     & \textbf{5.22} & \textbf{16.55} & 8.98 &\textbf{2.96} & 7.62  & \textbf{5.95}& 45.86 & 56.31 & \textbf{59.00} & \textbf{32.95} & \textbf{20.98} & \textbf{46.03}  \\   

SFL*    & 5.26  & 16.77& \textbf{8.95}& 3.02 & \textbf{7.42}  & 6.04     & \textbf{40.75}   & \textbf{31.06}    & 59.45   & 35.82   &  34.68  &  47.82 \\   \hline

STGCN           & 4.59    & 12.70    & 9.40    & 4.59    & 12.70    & 9.40     & 25.15   &   -      & 38.29   &  18.88  &   -     &  27.87  \\ 

Graph WaveNet   & 3.53    & 10.01    & 7.37    & 1.95    & 4.63     & 4.52     & 18.71   &  13.45   &  30.04  &  14.39  &  9.4    &  23.03   \\

\end{tabular}
}
\caption{Performance of traffic forecasting in federated setting}
\label{tb:main-result}
\end{table*}

\section{Experiment}
This section discusses the experimental analysis of the proposed method. In section \ref{sec-dataset}, we choose a graph-based benchmark dataset on traffic forecasting and an FL-based benchmark dataset on image classification. In section \ref{sec:baseline}, we introduce the baseline and experimental settings. Then, we compare the proposed SFL with other baselines and also perform the ablation study to verify this observation in section \ref{sec:comparison}. Further analysis is provided with visualization in section \ref{sec:visualization} and compatibility analysis in section \ref{sec:compatibility}. All implementation codes are available on Github\footnote{https://github.com/dawenzi098/SFL-Structural-Federated-Learning}.

\subsection{Datasets}\label{sec-dataset} The traffic datasets are ideal for validating our hypothesis, as the data comes with a natural topological structure and per-user data with non-IID distribution, all collected in the real world. We used four traffic datasets, METR-LA, PEMS-BAY, PEMS-D4, and PEMS-D8 to observe the performance of the SFL in different real-world scenarios. We apply the same data pre-processing procedures as described in \cite{wu2019graph}. All the readings are arranged in units of 5-minutes. The adjacency matrix is generated based on Gaussian kernel \cite{shuman2013emerging}. We also apply Z-score normalization to the inputs and separate the training-set, validation-set, and test-set in a 70\% 20\% and 10\% ratio. The evaluation metrics we use for the datasets include mean absolute error (MAE), root mean squared error(RMSE), and mean absolute percentage error (MAPE).

For the image datasets, we use benchmark datasets with the same train/test splits as in previous works \cite{krizhevsky2009learning}. We artificially partitioned the CIFAR-10 with parameter $k(shards)$ to control the level of non-IID data. The whole dataset is sorted according to the label and then split into $n \times k$ shards equally, and each of $n$ clients is assigned $k$ shards. In short, the smaller the $shards$ is, the more serious are the non-IID data issues.

\subsection{Baselines and experiment settings}\label{sec:baseline} We compare our method with four representative federated-learning frameworks including the standard FedAvg \cite{mcmahan2017communication} and three  other personalization federated frameworks, FedAtt \cite{ji2019learning}, FedProx \cite{li2020federated} and Scaffold \cite{karimireddy2020scaffold}. In addition, we also implement two fine-tune-based methods FedPer\cite{arivazhagan2019federated} and LG-FedAvg\cite{liang2020think}. During the client model selection, to focus more attention on the impact of introducing structural information during the server aggregation process, we choose simple and fixed client models for all frameworks to shield the influence of client model architecture. We use pure RNN for traffic prediction tasks with 64 hidden layer sizes. For CIFAR-10, we use ResNet9 as the base model for all evaluated methods, For a fair comparison, without any additional statement, all reported results are based on the same training set as follows. We employ SGD with the same learning rate as the optimizer for all training operations, use 128 for batch size, and the number of total communication rounded to 20. It is worth mentioning that higher capacity models and larger communication rounds can always produce a higher performance on any of those datasets. As such, the goal of our experiment is to compare the relative performances of these frameworks with the same basic models, rather than comparing specific values.

\begin{table*}[htpb]
\resizebox{2.0\columnwidth}{!}{
\begin{tabular}{|c|ccc|ccc|}
\hline
\multirow{2}{*}{} & \multicolumn{3}{c|}{Shards = 2} & \multicolumn{3}{c|}{Shards = 5} \\ \cline{2-7} 

&\multicolumn{1}{c|}{Mean Acc}  & \multicolumn{1}{c|}{Best 5\%} & Worst 5\% & 
\multicolumn{1}{c|}{Mean Acc} & \multicolumn{1}{c|}{Best 5\%} & Worst 5\% \\ \hline

FedAvg & \multicolumn{1}{c|}{18.55 ± 21.74}  & \multicolumn{1}{c|}{73.20 ± 10.93}  & 0.00 ± 0.00 &                        \multicolumn{1}{c|}{32.95 ± 17.61}  & \multicolumn{1}{c|}{67.40 ± 3.98}   & 2.20 ± 1.47   \\ \hline

FedAtt & \multicolumn{1}{c|}{10.08 ± 24.46}  & \multicolumn{1}{c|}{90.00± 20.00} & 0.00 ± 0.00         
       & \multicolumn{1}{c|}{28.25 ± 6.02}   & \multicolumn{1}{c|}{52.80 ± 0.75} & 1.40 ± 0.80     \\ \hline

FedProx  & \multicolumn{1}{c|}{12.49 ± 21.99} & \multicolumn{1}{c|}{74.20 ± 19.65} & 0.00 ± 0.00         
         & \multicolumn{1}{c|}{30.11 ± 14.85} & \multicolumn{1}{c|}{57.40 ± 1.85}  & 4.00 ± 1.90   \\ \hline

Scaffold & \multicolumn{1}{c|}{20.20 ± 26.73} & \multicolumn{1}{c|}{90.40 ± 1.85}  & 0.00 ± 0.00 
         & \multicolumn{1}{c|}{30.16 ± 13.66} & \multicolumn{1}{c|}{57.40  ± 5.57} & 2.40 ± 2.87   \\ \hline

FedPer & \multicolumn{1}{c|}{20.24 ± 18.52}   & \multicolumn{1}{c|}{78.30 ± 14.26} & 0.00 ± 0.00         
       & \multicolumn{1}{c|}{34.59 ± 18.26}   & \multicolumn{1}{c|}{69.25 ± 4.68}  & 4.29 ± 1.24   \\ \hline

LG-FedAvg & \multicolumn{1}{c|}{16.73 ± 22.01} & \multicolumn{1}{c|}{67.31 ± 12.68} & 0.00 ± 0.00         
          & \multicolumn{1}{c|}{31.75 ± 14.35} & \multicolumn{1}{c|}{67.24 ± 3.53}  & 2.73 ± 1.95  \\ \hline

SFL       & \multicolumn{1}{c|}{\textbf{54.25 ± 21.72}} & \multicolumn{1}{c|}{\textbf{100.00 ± 0.00}} & \textbf{6.2 ± 2.22} 
          & \multicolumn{1}{c|}{\textbf{45.03 ± 15.66}} & \multicolumn{1}{c|}{\textbf{75.20 ± 4.26}} & 
          9.20 ± 5.53   \\ \hline

SFL*     & \multicolumn{1}{c|}{50.54 ± 29.52}          & \multicolumn{1}{c|}{100.00 ± 0.00}          & 
0.00 ± 0.00         
         & \multicolumn{1}{c|}{36.18 ± 12.74}          & \multicolumn{1}{c|}{62.60 ± 1.02}          & \textbf{12.20 ± 2.64} \\ \hline
\end{tabular}
}
\caption{Performance comparisons on severe non-IID scenarios with CIFAR-10}
\label{tb:image-result1}
\end{table*}

\begin{table*}[htpb]
\resizebox{2.0\columnwidth}{!}{
\begin{tabular}{|c|ccc|ccc|}
\hline

\multirow{2}{*}{}   & \multicolumn{3}{c|}{Shards = 10}  & \multicolumn{3}{c|}{Shards = 20}  \\ \cline{2-7} 
    
& \multicolumn{1}{c|}{Mean Acc}              & \multicolumn{1}{c|}{Best 5\%}             & Worst 5\%           
& \multicolumn{1}{c|}{Mean Acc}              & \multicolumn{1}{c|}{Best 5\%}             & Worst 5\%  \\ \hline

FedAvg   & \multicolumn{1}{c|}{46.33 ± 11.69}  & \multicolumn{1}{c|}{69.40 ± 5.00}   & 20.40 ± 5.54          
         & \multicolumn{1}{c|}{81.80 ± 4.38}   & \multicolumn{1}{c|}{89.60 ± 1.36}   & \textbf{71.20 ± 1.94}
         \\ \hline

FedAtt   & \multicolumn{1}{c|}{40.00 ± 8.94}  & \multicolumn{1}{c|}{52.00 ± 4.00}    & 12.16 ± 1.40           
         & \multicolumn{1}{c|}{76.09 ± 6.31}  & \multicolumn{1}{c|}{82.00 ± 2.76}    & 44.02 ± 1.02  \\ \hline

FedProx  & \multicolumn{1}{c|}{45.85 ± 11.55} & \multicolumn{1}{c|}{68.20 ± 1.94}    & 21.00 ± 3.22          
         & \multicolumn{1}{c|}{81.94 ± 4.64}  & \multicolumn{1}{c|}{89.40 ± 1.96}    & 69.40 ± 2.33   \\ \hline

Scaffold & \multicolumn{1}{c|}{45.49 ± 11.36} & \multicolumn{1}{c|}{67.40 ± 2.24} & 21.60 ± 2.87 
         & \multicolumn{1}{c|}{\textbf{82.00 ± 4.38}} & \multicolumn{1}{c|}{\textbf{90.00 ± 1.10}} & 70.80 ± 2.99      \\ \hline
     
FedPer & \multicolumn{1}{c|}{46.65 ± 10.71}    & \multicolumn{1}{c|}{72.73 ± 3.71}   & \textbf{31.95 ± 4.24}          
       & \multicolumn{1}{c|}{81.71 ± 3.71}     & \multicolumn{1}{c|}{88.95 ± 1.23}   & 69.98 ± 1.95  \\ \hline
     
LG-FedAvg & \multicolumn{1}{c|}{45.63 ± 12.22} & \multicolumn{1}{c|}{65.13 ± 3.21}   & 26.23 ± 3.17          
         & \multicolumn{1}{c|}{80.33 ± 3.20}   & \multicolumn{1}{c|}{87.03 ± 1.30}    & 69.26 ± 1.73  \\ \hline

SFL      & \multicolumn{1}{c|}{\textbf{51.79 ± 14.04}} & \multicolumn{1}{c|}{\textbf{78.80 ± 2.56}} & 23.00 ± 4.90
         & \multicolumn{1}{c|}{81.70 ± 4.70}   & \multicolumn{1}{c|}{89.60 ± 0.80} & 69.60 ± 1.62 \\ \hline

SFL*     & \multicolumn{1}{c|}{44.20 ± 11.85}   & \multicolumn{1}{c|}{67.40 ± 2.24}  & 20.00 ± 3.85          
         & \multicolumn{1}{c|}{81.25 ± 4.78}    & \multicolumn{1}{c|}{89.20 ± 0.98}  & 68.40 ± 2.50 \\ \hline
\end{tabular}
}

\caption{Performance comparisons on relatively mild non-IID scenarios with CIFAR-10}
\label{tb:image-result2}
\end{table*}

\subsection{Comparison analysis}\label{sec:comparison} The performance of SFL in a traffic forecasting task compared with other baselines is provided in Table \ref{tb:main-result}. We use SFL* to denote the SFL with structure learning enabled. In this table, we report the average MAE, MAPE, and RMSE across all the clients for 60 minutes (12-time steps) ahead of prediction. The full result can be explored in three parts. First, for METR-LA and PEMS-BAY there is a 25\% and 18\% performance improvement in terms of MAE. Since the two datasets have relatively more nodes and complex structural information (edges) as stated in Table \ref{tb:main-result}, the use of a graph convolutional network to introduce sufficient structural information into the server aggregation process could significantly improve the performance of the FL system. Even compared with privacy non-preserved, the overall performance of our proposed method is still very competitive. Second, the PEMS-D4 provides us with a very practical scenario where the structural information is missing and the SFL cannot directly benefit from this lack of structural information. In this case, the results prove that our structure self-learning module can learn in the absence of information, thus bringing a greater than 10\% performance gain. Finally, the PEMS-D8 dataset provides the performance of SFL with a worst-case scenario where clients are few and far between. Here, the relationships are fragile. The results confirm that the performance lower-bound of SFL remains slightly better than the traditional methods due to the natural data distribution skew. This trait of SFL was carefully examined and analyzed in the next set of experiments.

We then ran experiments on the image-based CIFAR-10 dataset to further validate SFL's ability to deal with the non-IID data. Table \ref{tb:image-result1} and \ref{tb:image-result2} demonstrates SFL's superior performance in different levels of distribution skew for non-IID. Note that the larger value of $shards$ indicates that the data is distributed more evenly across clients. For CIFAR-10, with the mimics of an severe non-IID data environment (shards are 2 and 5), the traditional FL algorithms are not functional. Our SFL performs significantly better than other algorithms, both from the best 5\% and worst 5\% due to its unique aggregation mechanism. As the data distribution tends to iid (shards are 10 and 20), the performance of the traditional algorithms increases to the normal level while our SFL still maintains a very competitive performance.

\begin{figure}[hbp!]
\vspace{-1em}
\centering
\includegraphics[width=\linewidth]{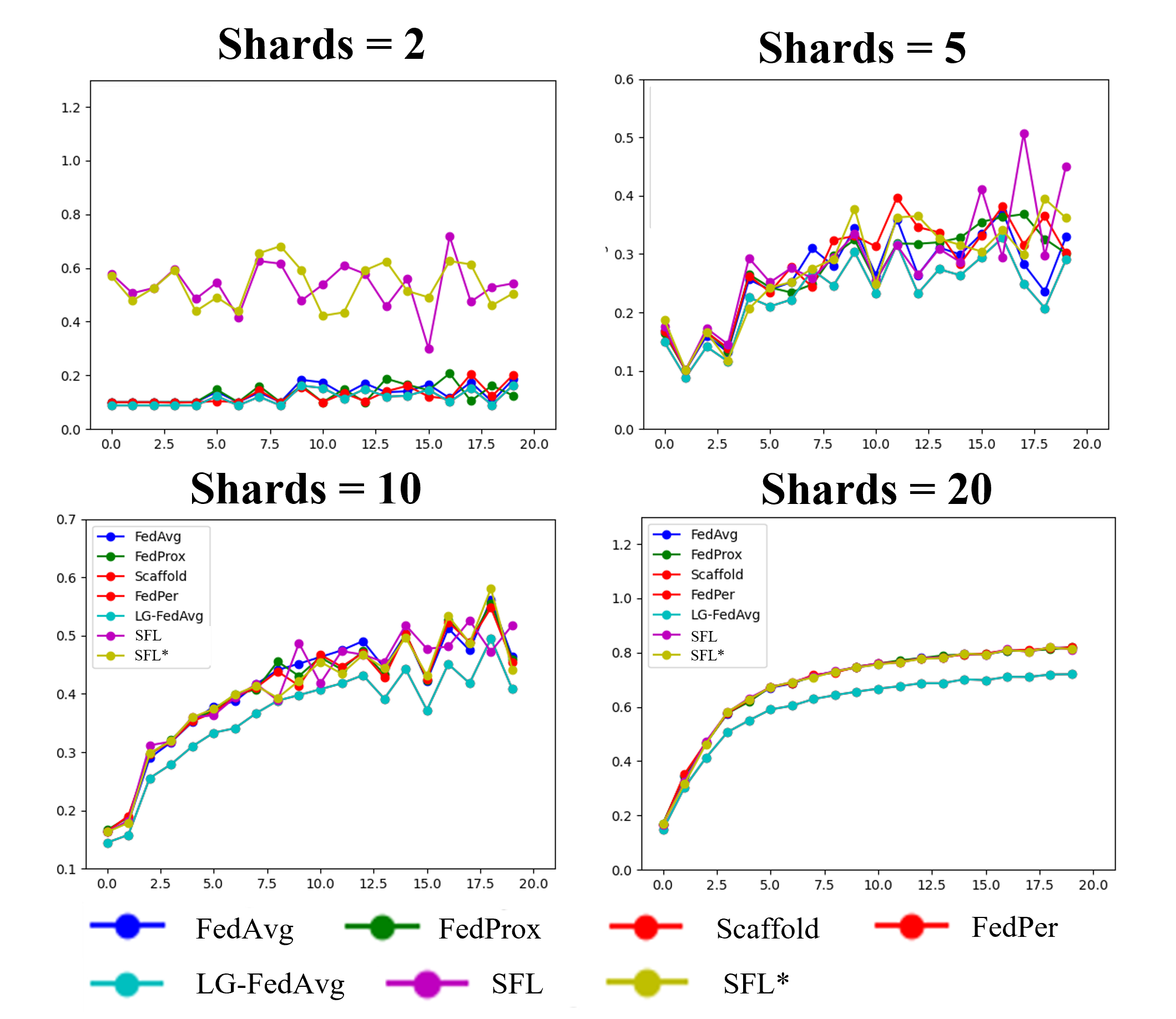}
\caption{Visualization of Convergence}
\vspace{-1em}
\label{fig:visul}
\end{figure}

\subsection{Visualization}\label{sec:visualization} Fig. \ref{fig:visul} illustrates the convergence process of SFL in different non-IID scenarios. Under the severe conditions where shards=2, there is only a small overlap in the client data distributions, which results in serious parameter conflicts during the server aggregation process, with all algorithms failing to converge. Using SFL can reduce the client parameters' conflict, thus producing a better result. When shards=5 or 10, SFL still has an obvious advantage on both convergence and robustness. When shards=20, most FL methods perform similarly because the data distribution has nearly become an IID scenarios.

\begin{figure}[h]
\centering
\includegraphics[width=\linewidth]{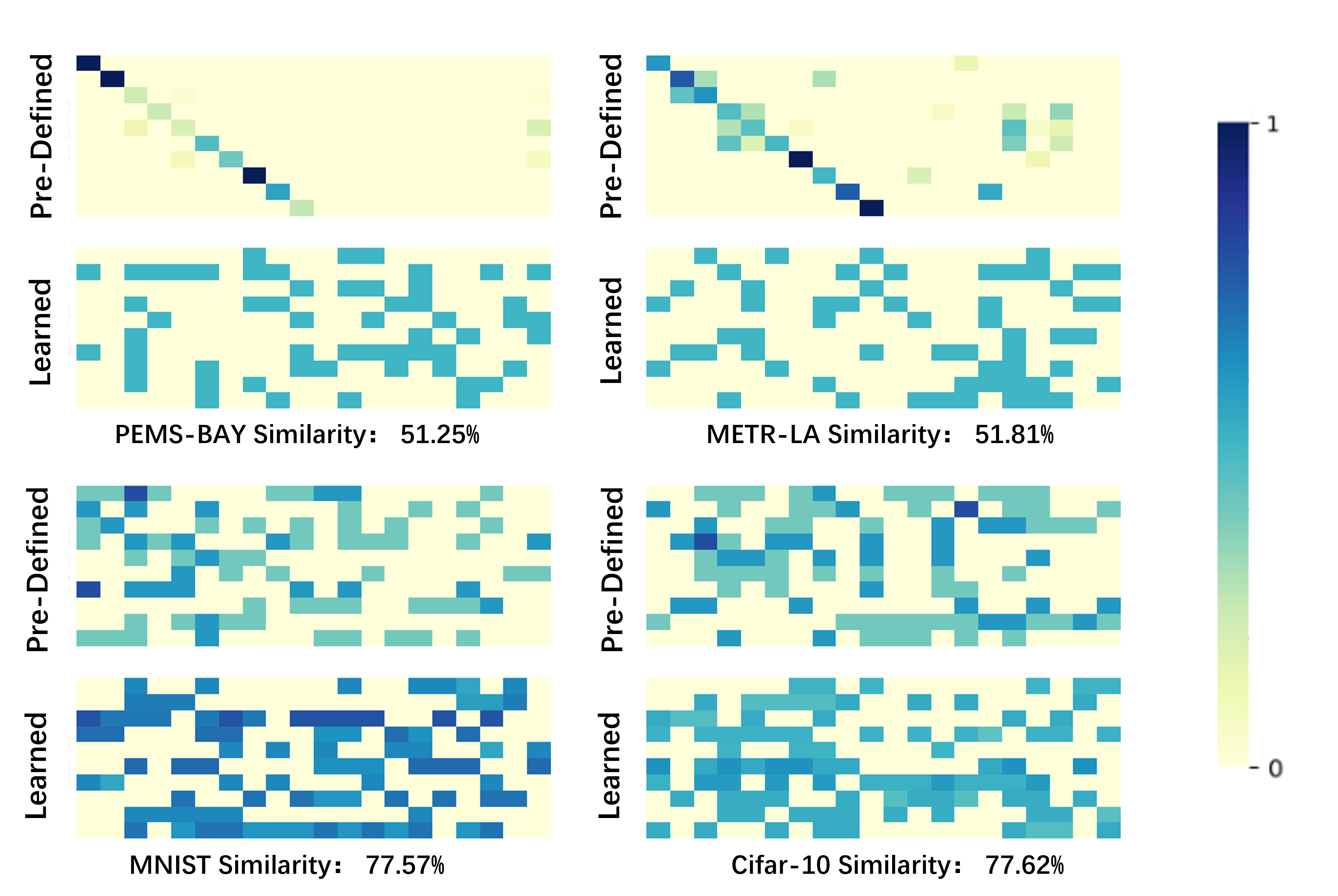}
\caption{Visualizing adjacent matrix comparison between pre-defined graph and learned graph on four datasets}
\setlength {\belowcaptionskip} {-1.5em}
\label{fig:visual-matrix}
\end{figure}

Fig. \ref{fig:visual-matrix} visualizes the comparison of the pre-defined relation graph and the learned graph by SFL. Twenty-five clients were selected to visualize their partial pre-defined structural information (top half) and the learned graph (bottom half). For image classification tasks on MNIST and CIFA-10, when sharks = 5, the learned adjacent connection approaches pre-defined adjacent values. In particular, the learned graph on traffic data is relatively "comprehensive" than the pre-defined graph because the pre-defined graphs of PEMS-LA and PEMS-BAY are purely constructed by a road connection relationship. In contrast, the learned graph includes the long-dependence on non-connected roads. This visualization demonstrates that SFL not only learns knowledge from simple pre-defined graphs but also learns comprehensive hidden relationships among clients.

\begin{table}[h]
\centering
\begin{tabular}{|c|c|c|c|}
\hline
                 &  SFL   & SFL+LG         & SFL+PER       \\ \hline
CIFAR-10(Acc \%) &  45.03 & \textbf{46.32} & 45.93         \\ \hline
PEMS-BAY(MAE)    &  6.47  & 4.95           & \textbf{4.82} \\ \hline
\end{tabular}
\caption{Compatibility Performance}
\label{tb:compatibility}
\end{table}

\subsection{Compatibility analysis} \label{sec:compatibility} SFL can be combined with the existing PFL methods to further improve the performance. By integrating SFL with another two PFL methods, we train the PEMS-BAY and CIFAR-10 datasets (shards=5) for 20 communication rounds. Specifically, instead of applying personalized fine-tuning based on the shared global model from FedAvg, we applied two types of personalization processes: LG-FedAvg \cite{liang2020think} and FedPER \cite{arivazhagan2019federated} on top of the SFL. As stated in Table \ref{tb:compatibility}, SFL can combine with existing PFL methods to further improve the performance of FL.

\section{Conclusion}
This paper is the first work to study graph-guided personalized FL. A GCN-based model aggregation mechanism has been adopted to facilitate implementation. SFL can not only leverage the graph structure but and also discover comprehensive hidden relationships amongst clients. Experiments on GNN-based traffic dataset and FL-based image benchmark datasets have demonstrated the effectiveness of SFL.

\newpage
\newpage
\bibliographystyle{named}
\bibliography{ijcai22}

\end{document}